\documentclass[letterpaper, 10 pt, conference]{ieeeconf}  % Comment this line out if you need a4paper
\usepackage{graphicx}
\usepackage{url}
\usepackage[ruled, linesnumbered, noend]{algorithm2e}

\makeatletter
\renewcommand*{\@algocf@post@ruled}{}
\makeatother

\usepackage[margin=0pt,font=small,labelfont=bf]{caption}
\usepackage{subcaption}

\usepackage{multirow}
\usepackage{amsfonts}
\usepackage[most]{tcolorbox}

\usepackage[normalem]{ulem}
\usepackage{todonotes}
\usepackage{booktabs}
\usepackage{cite}
\usepackage{amssymb}
\usepackage[a-1b]{pdfx}

\def\dch{\textcolor{black}}

\def\dadd{\textcolor{black}}
\def\ddel#1{{}}
\def\dmod{\textcolor{black}}
\def\daddblack{\textcolor{black}}
\def\dmodblack{\textcolor{black}}

\IEEEoverridecommandlockouts                              % This command is only needed if 
\title{\LARGE \bf
On-the-Go Tree Detection and Geometric Traits Estimation \\with Ground Mobile Robots in Fruit Tree Groves
}

\author{Dimitrios Chatziparaschis,$^{\dagger}$ Hanzhe Teng,$^{\dagger}$ Yipeng Wang,$^{\dagger}$ Pamodya Peiris,$^{\dagger}$ \\Elia Scudiero,$^{\ddagger}$ and Konstantinos Karydis$^{\dagger}$% <-this % stops a space
\thanks{$^{\dagger}$~Dept. of Electrical and Computer Engineering, and~$^{\ddagger}$~Dept. of Environmental Sciences, Univ. of California, Riverside, 900 University Avenue, Riverside, CA 92521, USA;
{\tt\footnotesize\{dchat013, hteng007, ywang1040, ppeir022,  scudiero, karydis\}@ucr.edu}.}%
\thanks{We gratefully acknowledge the support of NSF \# IIS-1901379 and \# CMMI-2046270, USDA-NIFA \# 2021-67022-33453, ONR \# N00014-18-1-2252 and \# N00014-19-1-2264, The University of California \# UC-MRPI M21PR3417, a Frank G. and Janice B. Delfino Agricultural Technology Research Initiative Seed award and an OASIS-IFA award. 
Any opinions, findings, and conclusions or recommendations expressed in this material are those of the authors and do not necessarily reflect the views of the funding agencies.}% <-this % stops a space%
}

\usepackage[switch, modulo]{lineno}
\SetKwHangingKw{kwContinue}{continue}
\SetKwHangingKw{kwReceive}{retrieve}
\SetKwHangingKw{kwExecute}{execute}
\SetKwHangingKw{kwFind}{find}
\SetKwHangingKw{kwDefine}{define}
\SetKwHangingKw{kwCompute}{compute}
\SetKwHangingKw{kwReturn}{return}

\begin{document}

\maketitle
\thispagestyle{empty}
\pagestyle{empty}

%%%%%%%%%%%%%%%%%%%%%%%%%%%%%%%%%%%%%%%%%%%%%%%%%%%%%%%%%%%%%%%%%%%%%%%%%%%%%%%%
\begin{abstract}

By-tree information gathering is an essential task in precision agriculture achieved by ground mobile sensors, but it can be time- and labor-intensive. In this paper we present an algorithmic framework to perform real-time and on-the-go detection of trees and key geometric characteristics (namely, width and height) with wheeled mobile robots in the field. Our method is based on the fusion of 2D domain-specific data (normalized difference vegetation index [NDVI] acquired via a red-green-near-infrared [RGN] camera) and 3D LiDAR point clouds, via a customized tree landmark association and parameter estimation algorithm. The proposed system features a multi-modal and entropy-based landmark correspondences approach, integrated into an underlying Kalman filter system to recognize the surrounding trees and jointly estimate their spatial and vegetation-based characteristics. Realistic simulated tests are used to evaluate our proposed algorithm's behavior in a variety of settings. Physical experiments in agricultural fields help validate our method's efficacy in acquiring accurate by-tree information on-the-go and in real-time by employing only onboard computational and sensing resources.

\end{abstract}

% \listoftodos

% including .tex submodules
\section{Introduction}\label{seq:introduction}

Real-time tree crop monitoring is important in agriculture because it allows growers to make informed decisions about management practices, address potential issues promptly, and ultimately optimize crop yields and profitability. Growth rates, yields, and fruit quality can vary significantly on a tree-by-tree basis in commercial orchards due to the spatiotemporal variability of environmental factors such as soil conditions and microclimate~\cite{perry2010spatial}, as well as management practices (e.g., non-uniform irrigation distribution~\cite{mcclymont2011variation}) and their interaction with individual tree genetic variability~\cite{coupel2019multi}. Unfortunately, it is impractical to directly measure the status (e.g., leaf/stem water potential, fruit quantity and quality, etc.) over time and for every single tree. Thus, different types of technology have been employed to gather indirect (i.e. proximal and remote sensing) by-tree data in an efficient and scalable manner. 

%9865266
Unmanned Aerial Vehicles (UAVs), in particular, have been deployed in farms to acquire information from above and map field characteristics~\cite{9705188,8782102, chatziparaschis2023real, RADOGLOUGRAMMATIKIS2020107148}. Yet, the accuracy of these approaches is highly connected to the resolution of the used sensors and flight altitude. Moreover, UAVs are not practical for collecting data from under or on the sides of tree canopies. 
More recently, mobile robots have been used for in-field near-ground sensing to collect various types of data about tree geometric traits and health markers~\cite{7059344, 8066090}.

Of interest to this work is in-situ proximal sensing~\cite{doi:https://doi.org/10.2134/precisionagbasics.2016.0093} with ground robots. 
In our previous works we have considered proximal sensing of soil apparent electrical conductivity which can help estimate spatial variability of many agronomically-relevant soil properties, such as soil moisture, particle size fraction, and salinity~\cite{campbell, chatziparaschis2023robotassisted, elia2023microwaveradio}, and developed a leaf visual detection and pose estimation system which can assist toward plant phenotyping~\cite{campbell2022integrated,dechemi2023watering}. 
We envision that through data fusion from multiple ground (and remote) sensors, we may apply current automated sampling methods~\cite{kan2021stocgrahs} and develop new models to assess orchard status on-the-go.

In this work, we develop a real-time algorithmic framework for tree-crop proximal sensing of plant geometric and vegetation index markers by using a semi-autonomous ground mobile robot in large-scale field environments. 
Our robot %(Fig.~\ref{fig:jackal_system_in_grove}) 
is equipped with a 3D LiDAR sensor and an RGN (Red-Green-NIR) camera module, and performs real-time fusion of the two sensor modalities to obtain and identify the surrounding tree candidates and extract their geometric and vegetation index information. 
We present a novel landmark association approach based on both 2D and 3D sensing modalities to obtain real-time tree correlations and match them with globally georeferenced trees to update their status in the surveying map.  
In sum, this work contributes:
\begin{itemize}
    \item A real-time LiDAR-camera fusion algorithm to obtain accurately the surrounding tree geometric traits and geo-reference attained information on a global map. 
    \item A multivariate, entropy-based correspondences algorithm, integrated into an underlying Kalman Filter, for robust 3D tree cluster matching for localization. 
    \item Robot-assisted proximal RGN sensing from the ground enabling integration of domain-specific vegetation data for active landmark detection outdoors.
\end{itemize}

Our framework is deployed and tested experimentally in the field under real-world conditions (e.g., significant wind gusts and frequent ambient light variations). 
Algorithm efficacy is also tested within realistic digital twins over multiple distinctive case studies. 
In both settings, we compare our system with a conceptually-related tree detection methodology and present the results. 
Our whole system is autonomous-ready, e.g., by employing Global Navigation Satellite System (GNSS) and waypoint navigation, to help stimulate further agricultural robotics research, but it can be integrated directly into existing commercial (semi-) autonomous agricultural vehicles (e.g., smart tractors) as well. 
\vspace{-3pt}

\section{Related Works}\label{seq:relatedwork}

Tree detection and by-tree geometric traits estimation have been active research areas in both remote sensing and robotics. Typically (airborne) data are collected and then post-processed (e.g.,~\cite{4378541,9645363,Vauhkonen2011ComparativeTO,4156171}) to obtain counts of observed trees, crop density and growth information, and field properties (e.g., soil moisture). Machine learning approaches have been employed as well~\cite{9800966, WAN2022106609, 8930566, 8865635} on imagery and point cloud data.
Using offboard processing, Zhang $et~al.$~\cite{rs11020211} developed a method for standing tree stem detection in rich LiDAR data based on the point cloud's surface curvature and a segmentation-based approach. 
Also, Roy $et~al.$ \cite{8594167} demonstrated an offline method to merge reconstructed views from two sides of an orchard row into a unified 3D model.

In studies involving robot-assisted proximal sensing, trees are often used as landmarks for localization. In most cases, tree detection relies on trunk recognition, as the trunk serves as a sturdy landmark in dynamic environments. Jelavic~$et~al.$~\cite{Jelavic2021RoboticPH} showcased a robust and lightweight method for tree detection based on Principal Component Analysis (PCA) on the tree clusters, given 3D LiDAR scans in local forest patches. 
Shalal~$et~al.$~\cite{SHALAL2015254} demonstrated a mobile robot with an integrated 2D LiDAR sensor and a color camera, and  Yu~$et~al.$~\cite{9517543} presented a localization system based on fusion of camera and ultrasonic data.
In the same context, 
Durand-Petiteville~$et~al.$~\cite{8412556} 
used four stereo cameras to perceive tree trunks through the formed concavities in the point cloud space. 
Although these studies demonstrated practical applications in tree detection, it is notable that trunks may not always be directly visible in the scene, as the latter can be drastically affected by dense, dynamic, and verdant environmental conditions.  

To this end, several works have proposed using trained detectors for trunk (or other tree parts) to improve robustness. 
Wang~$et~al.$~\cite{9562056} presented a PointRCC detector that is trained on stereo data (pseudo-lidar representation) to identify trees in a forest environment. Liu~$et~al.$~\cite{8653965} demonstrated a monocular camera and a LiDAR system for fruit counting and tree trunk detection method, based on Convolutional Neural Networks. Also, Semantic LiDAR Odometry and Mapping (SLOAM)~\cite{ctx48528145780003681} has an integrated tree detection algorithm based on trunk detection, which is trained on a custom dataset created in forest environments. 
However, the performance of these approaches highly depends on additional computational resources, precise training, and the generalizability of the method to be robust in different environments and physical conditions of the trees.

To the author's best of knowledge, there is a lack of works on full tree characteristics estimation from the ground utilizing vegetation-enabled features. 
We present our approach to localizing tree landmarks, associating them with existing tree candidates, and spatially characterizing them without exclusively relying on any specific part of their structure. 
We also test against the most relevant system~\cite{Jelavic2021RoboticPH}, since it can obtain full tree-cluster dimensions during the navigation of a robotic harvester in a forest.

\vspace{-2pt}

\section{System Description}\label{seq:systemdescription}

\begin{figure}[!t]
\vspace{6pt}
     \centering
     \begin{subfigure}{.25\textwidth}
         \centering
  \includegraphics[trim={0 5cm 0 16cm},clip,height=2cm]{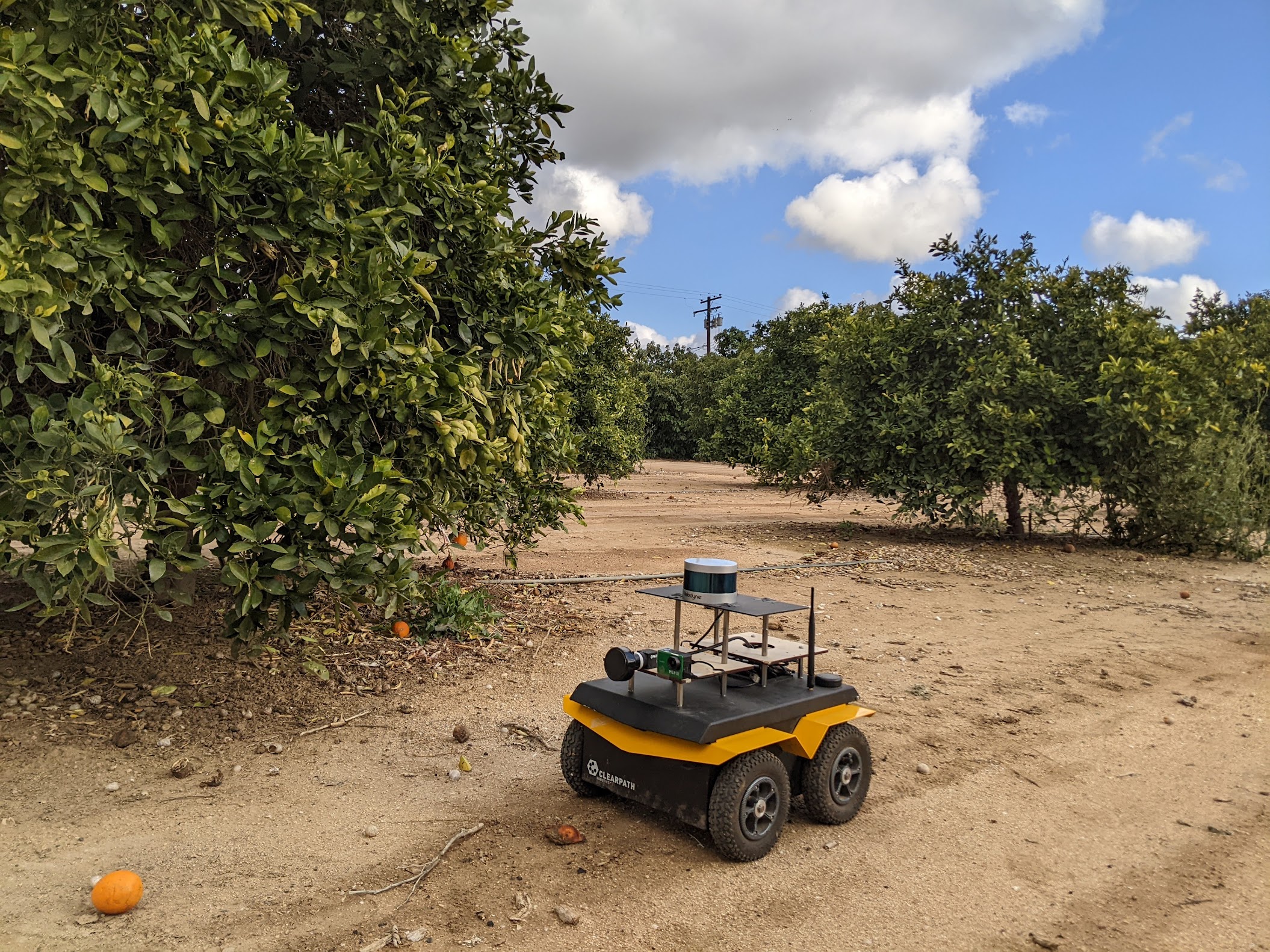}
  \vspace{-4pt}
         \caption{}
         \label{fig:jackal_system_in_grove}
     \end{subfigure}\hfill
     \begin{subfigure}{.2\textwidth}
         \centering
    \includegraphics[trim={0 1cm 0 1cm},clip,height=2cm]{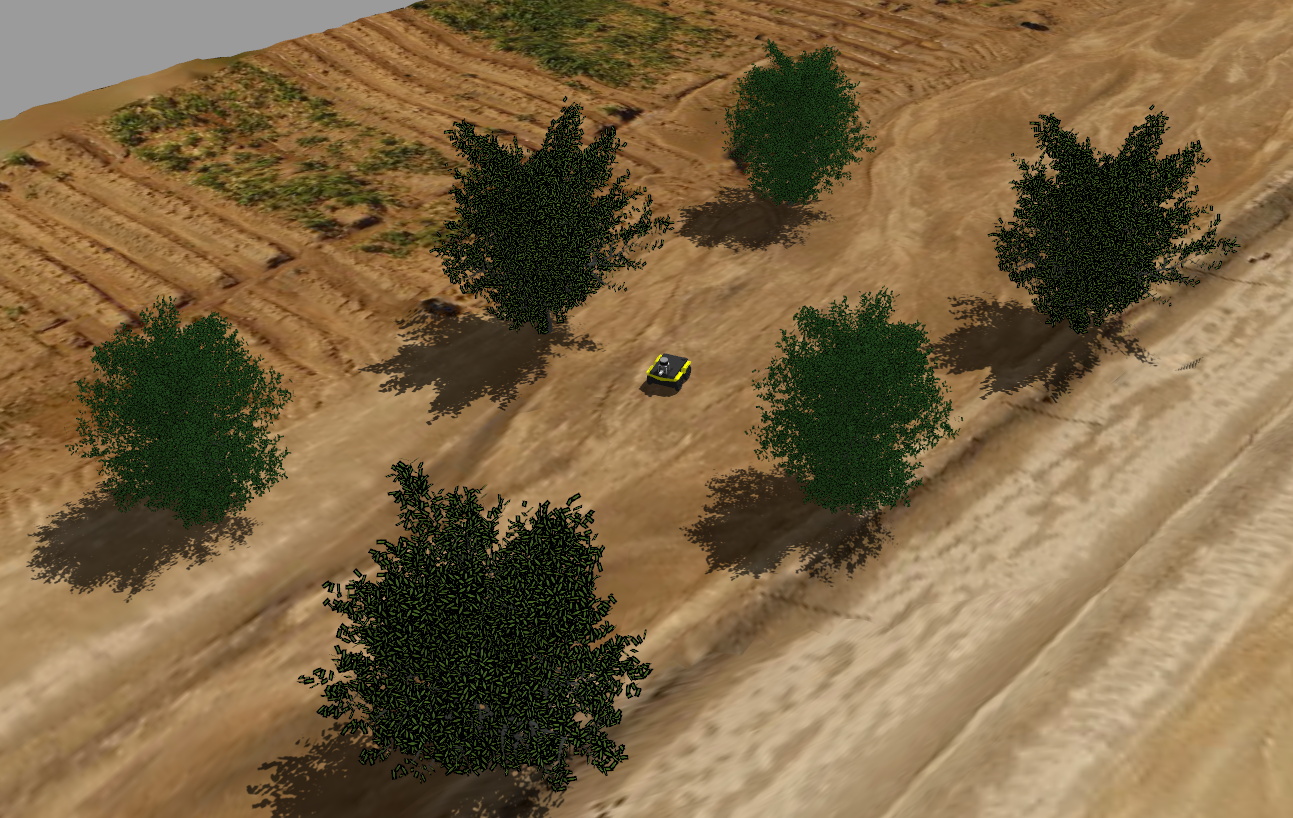}
  \vspace{-4pt}
         \caption{}
         \label{fig:gazebo_simulated_world}
     \end{subfigure}
     \vspace{-4pt}
    \caption{(a) Clearpath Jackal equipped with a Velodyne VLP-16 LiDAR and a MAPIR Survey3N RGN camera during field experiments. (b) The designed simulation world in Gazebo along with the simulated Jackal mobile robot. This world model contains a $2\times 3$ tree grid of 3 orange and 3 lemon trees placed at a $5~m$ width $\times$ $5~m$ length relative distance from each other.  
    }\label{fig:blank}
    \vspace{-12pt}
\end{figure}

\begin{figure}[!t]
\vspace{6pt}
     \centering
     \begin{subfigure}{.23\textwidth}
         \centering
  \includegraphics[width=4cm, trim={0 0 0 0.1cm},clip]{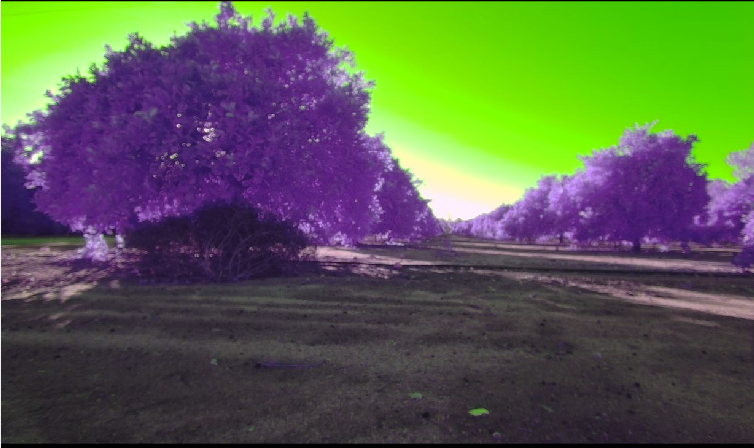}
  \vspace{-15pt}
         \caption{}
         \label{fig:rgn_raw}
     \end{subfigure}\hfill
     \begin{subfigure}{.23\textwidth}
         \centering
  \includegraphics[width=4cm, trim={0 0.15cm 0 0},clip]{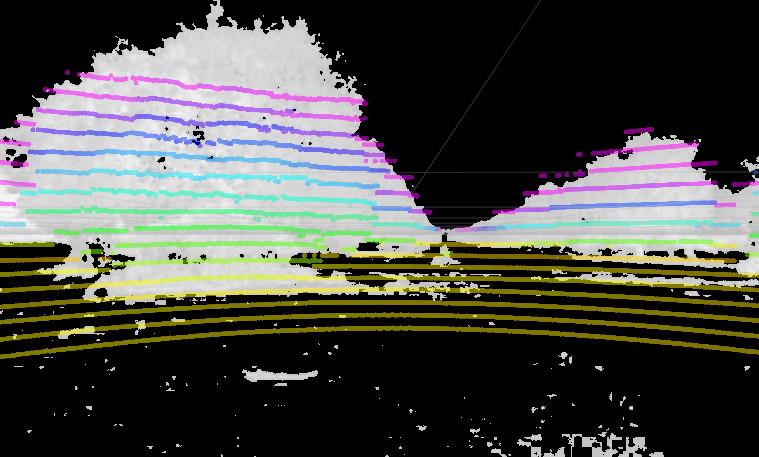}
   \vspace{-15pt}
         \caption{}
         \label{fig:ndvi_fused}
     \end{subfigure}
     \vspace{-3pt}
    \caption{Snapshot from robot operation in the orchard. (a) RGN raw imagery data. (b) Corresponding binary thresholded NDVI frame with projected 3D LiDAR data colorized based on $z$-level value.}
    \label{fig:rgn_frames}
    \vspace{-19pt}
\end{figure}

We use the Clearpath Jackal wheeled mobile robot (Figure~\ref{fig:jackal_system_in_grove}).\footnote{~We note that the algorithm developed herein is general and it can be employed with other wheeled robots that can operate in the field too.} The robot has body dimensions of $ 50.8~cm \times 43.2~cm \times 25.4~cm $ and weighs $17~kg$. It features an NVIDIA Xavier AGX internal PC, a GPS receiver, an integrated IMU sensor, and two wheel encoders. 
It is equipped with a 3D LiDAR sensor (Velodyne VLP-16) and a RGN camera (MAPIR Survey3N); both are affixed to the robot's chassis.

\paragraph*{\bf 3D LiDAR} The publication rate of the LiDAR sensor was set to $10Hz$ thus providing up to $300,000$ points per second. 
Each point in the pointcloud $\mathcal{P}\subseteq \mathbb{R}^3$ holds also information about the 3D LiDAR $ring\_id$ that was detected.

\paragraph*{\bf RGN Camera} The main purpose of the RGN camera is the inclusion of Near Infrared (NIR) band in imaging. 
Given readings in the red and green channels of the sensor, NIR provides critical information used to compute the Normalized Difference Vegetation Index (NDVI), %as well as the Green NDVI (GNDVI). These are 
defined as $NDVI = (NIR-Red)/(NIR+Red)$.
NDVI can help pinpoint and quantify areas with lower chlorophyll status and vegetation health, since it is a ratio of near-infrared radiation (highly reflected by vegetation) and red light (mainly absorbed by vegetation)~\cite{phenotypictechniques}.

The camera provides $30$\;fps of $720$p RGN feed through a USB connection directly to the onboard computer. Figure \ref{fig:rgn_raw} shows a snapshot of the raw RGN imagery information captured in the field. 
During field survey, we use the reflectance calibration ground target package (V2) to regulate the minimum and maximum reflectance values and obtain NDVI %and GNDVI 
readings in the low-high normalized health range of $[-1.0,1.0]$, respectively. 
We also calibrated the camera to acquire intrinsic properties and lens distortion coefficients~\cite{zhang2000flexible}, used for 3D LiDAR points' projection on the RGN image plane. Let $q_\mathcal{O}\in\mathcal{O}\subseteq\mathbb{R}^3$ be the 3D points in the camera's coordinate system and $q_{x,y}\in\Pi\subseteq\mathbb{R}^2$ their representation on the image plane on pixel position $(x,y)$.

\paragraph*{\bf LiDAR -- RGN Camera Extrinsic Calibration}

Key in our LiDAR-camera fused detection approach (Section~\ref{seq:approach}) is to define the relative pose of the two sensors (i.e. extrinsic calibration). 
To do so, we employed the package developed in~\cite{DBLP:journals/corr/abs-2103-12287}, which is a scene-based calibration approach to estimate the extrinsic parameters and obtain the LiDAR-to-camera frame transformation matrix $_{\mathcal{P}}^{\Pi}T\in SE(\mathit{3})$. 
Figure~\ref{fig:ndvi_fused} shows projected LiDAR points on the NDVI frame.

\paragraph*{\bf Positioning in Global and Local Frames}
To correlate detected tree candidates in the environment with ground truth tree landmarks, we employ the robot localization software~\cite{ekf2} from the ROS~\cite{quigley2009ros} navigation stack. 
It uses the robot's GNSS data, the earth-referenced heading, and odometry information to generate a Cartesian datum in the Universal Transverse Mercator (UTM) coordinate system. 
This aligns with our experiments (Section~\ref{seq:experiments}) as we assume access to tree field geomaps and the robot's initial pose.

\section{Proposed Algorithm}\label{seq:approach}

\begin{figure}[!t]
    \vspace{0pt}
    \centering
    \includegraphics[width=7.25cm]{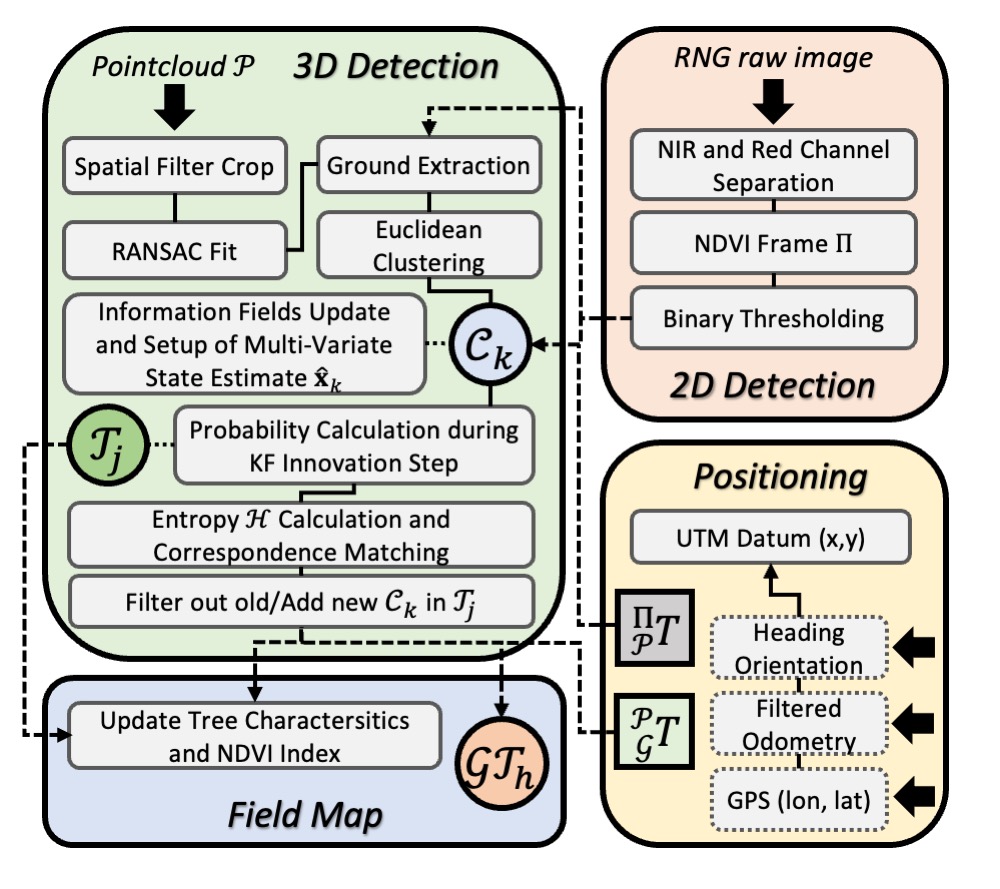}
    \vspace{-10pt}
    \caption{System diagram of the proposed 2D-3D fusion system with main sub-components, along with the positioning node, the field map node, and the raw sensory input streams. The solid lines indicate the inner connectivity of the components and the dotted arrows the information exchange between the developed nodes.} 
    \label{fig:system_sketch}
    \vspace{-20pt}
\end{figure}

Figure~\ref{fig:system_sketch} illustrates the overall flow diagram of our framework. 
The main parts of the framework include multi-modal data fusion, correspondence matching, and estimation of desired traits (herein tree height and width). 

\begin{algorithm}
\vspace{1pt}
\caption{Tree Detection, Identification, and Characteristics Estimation}
\label{alg:treedetectionalgorithm}
  \scriptsize
    \SetKwInOut{Input}{\hspace{0.42cm}input}
  \SetKwInOut{Output}{\hspace{0.42cm}output}
  \Indm\Indmm
  % \Input{$vox_{conf}$, $eucl_{conf}$, $min\_vol\_thres$, $low\_prob\_thres$, $t_{time\_thres}$}
  \Input{$key\_params$, $v_x, \omega_{\theta}$}
  \Output{\daddblack{Global georeferenced} trees $\mathcal{GT}$ with their characteristics}
\Indp\Indpp
  \kwDefine{$\mathcal{T}=\dmodblack{\varnothing}$ tree candidates/landmarks in memory}
  \While{newly received \dmodblack{point cloud} $\mathcal{P}$}{
  \vspace{0.1cm}
    $\dmodblack{\mathcal{C}_k} = compute\_clusters(\dmodblack{\mathcal{P}});$\\
      \vspace{0.1cm}
    \For{\dmodblack{all clusters $k$ in $ \mathcal{C} \subseteq \mathcal{P}$}}{
      \vspace{0.1cm}
     $m_k = compute\_characteristics(k,key\_params);$\\
       \vspace{0.1cm}
     \If{$\mathcal{T}=\dmodblack{\varnothing}$}{
        $\dmodblack{\hat{\mathbf{x}}_k} = m_k;~\mathbf{P}_k = I_{\dmodblack{length}(key\_params)}$;\\
        \kwDefine{$new\_tree_k:~\dmodblack{\hat{\mathbf{x}}_k}, \mathbf{P}_k,\{\mathbf{F}_k,\mathbf{H}_k,\mathbf{Q}_k,\mathbf{R}_k\}$}%\vspace{-0.4cm}
        \kwExecute{$\mathcal{T}.insert(new\_tree_{k});~\textbf{\dmodblack{continue}}$}
 }
  \vspace{0.1cm}
\For{all trees $j$ in $\mathcal{T}$}{
    $\hat{\mathbf{x}}_j \leftarrow \mathbf{F}_j \dmodblack{\hat{\mathbf{x}}_j} + \mathbf{B}_j v_x;$\\
    $\mathbf{i}_j = norm\_residuals(m_k - \mathbf{H}_j\hat{\mathbf{x}}_j);$\\
    $\mathbf{S}_{j}=\mathbf{H}_{j}{\mathbf{P}}_{j}\mathbf{H} _{j}^{\textsf{T}}+\mathbf{R} _{j};$\\
    $^{\boldsymbol{k}}\boldsymbol{p}_j = exp\big( -0.1\cdot d_{\mathcal{M}}(\mathbf{i}_j,\mathbf{S}_{j}^{-1})\big) \in [0,1];$
  }
  \vspace{0.1cm}
  $max_j = \underset{j}{argmax}\big(\dmodblack{^{\boldsymbol{k}}\boldsymbol{p}_j\big)};$ \\
  \vspace{0.1cm}
  $\mathcal{H} = - \sum\limits_{j} \dmodblack{^{\boldsymbol{k}}\boldsymbol{\hat{p}}_j} log_{\dmodblack{M}}(\dmodblack{^{\boldsymbol{k}}\boldsymbol{\hat{p}}_j})~where~\sum\limits_{j}\dmodblack{^{\boldsymbol{k}}\boldsymbol{\hat{p}}_j}=1; $\\
  \vspace{0.1cm}
    \eIf{$^{\boldsymbol{k}}\boldsymbol{p}_{max_j}>th_p~and~\mathcal{H}<th_\mathcal{H}$}{
        \kwExecute{$association\_pairs.insert\Big([\mathcal{T}_{max_j}, new\_tree_{k}]\Big)$}
  \vspace{-0.4cm}
        } {
        \kwExecute{$\mathcal{T}.insert(new\_tree_{k})$}         
     }
  }
  \vspace{0.1cm}
    \For{\dmodblack{all $association\_pairs \sim \Big(\forall~[\mathcal{T}_{j'},new\_tree_{k'}]\Big) $}}{
        
        $\dmodblack{Kalman\_Update}(m_{k'},\dmodblack{\hat{\mathbf{x}} _{j'}},\mathbf {P}_{j'});$\\
        $\mathcal{T}_{j'} = new\_tree_{k'};$
    }
  \vspace{0.2cm}
 \For{\dmodblack{all trees $j$ in $\mathcal{T}$}}{
    \kwExecute{$\dmodblack{min_h}=\underset{h}{argmin}\Big(\dmodblack{\ell^2}\mbox{-}norm(\mathcal{GT}_h~,~_{\mathcal{P}}^{\mathcal{G}}T^{-1} \cdot\mathcal{T}_j)\Big)$}
    \kwExecute{$\mathcal{GT}_{\dmodblack{min_h}} = \mathcal{T}_j$}
    }
}
\vspace{-20pt}
\end{algorithm}

\subsection{2D and 3D Data Fusion and Tree Detection}\label{ssec:fusion}

The overall goal of our algorithm is to detect and recognize $M$ tree candidates, $\mathcal{T}_j$ with $j\in[1,M]$, in a captured point cloud $\mathcal{P}$, and match them with $N$ global georeferenced trees on the field, $\mathcal{GT}_h$ with $h\in[1,N]$, to update their width and height information in real-time and on-the-go.

First, the three channels in the RGN camera are split to compute the points' NDVI value $v_{ndvi}\in\mathbb{R}$. 
Binary thresholding is applied to the computed single-channel NDVI frame to segment the vegetation areas (selected values were set in the range $q_{x,y}\in(-0.3,1.0]$ following calibration testing in the field). 
At the same time, spatial filtering and Random Sample Consensus (RANSAC) fit is applied to the received point cloud $\mathcal{P}$ along with an additional check of point projection on $\Pi$ to remove points that belong on the ground surface. 
The remaining point cloud is downsampled and Euclidean Clustering~\cite{rusu2010semantic} is applied to discretize and obtain appearing clusters in the $\mathcal{P}$ space, all while preserving the closest $v_{ndvi}$ values. 
The process takes place in the function $compute\_clusters$ in line 3 of Algorithm~\ref{alg:treedetectionalgorithm}.

For each cluster $\mathcal{C}_k=\{p \in \mathcal{P}; v_{ndvi}\}\in\mathbb{R}^4$ we compute the centroid point $p_{ct}\in\mathbb{R}^3$, the total number of points $num_{pt}\in\mathbb{N}$, and the average NDVI value $\bar{v}_{ndvi}\in\mathbb{R}$ as the main correspondence matching characteristics. 
These features will be used for the state estimation of each landmark $\mathcal{T}_j$.\footnote{~It is possible to consider additional attributes, e.g., convex hull area and volume of the points included in each $\mathcal{C}_k$. We anticipate that doing so will have limited impact on the pipeline's runtime considering the small $\mathcal{C}_k$ size while increasing system capabilities. A more formal treatise of this topic falls outside the scope of this paper, and is part of future work.}
We thus have $\hat{\mathbf{x}}_k=[p_{ct}, num_{pt}, \bar{v}_{ndvi}]\in\mathbb{R}^{5\times 1}$ and $\mathbf{P}_k=I_5$, where $\hat{\mathbf{x}}_k$ is the state estimate and $\mathbf{P}_k$ the estimate covariance. 
Lines 6 to 9, indicate the initialization of the first tree clusters in memory, as individual Kalman Filter estimates with the internal models $\{\mathbf{F}_k,\mathbf{H}_k,\mathbf{Q}_k\}=I_5$ and $\mathbf{R}_k=0.1\cdot I_5$.

\iffalse
\begin{table}[!h]
\vspace{-6pt}
\begin{center}
\footnotesize
\caption{\footnotesize Key Parameters of Tree Clusters \dmod{$\mathcal{C}_k$} and Candidates $\mathcal{T}_j$ \label{tab:informationfields}}
\vspace{-5pt}
\begin{tabular}{cl} \toprule
 Parameter name   &Information Field \\ \midrule
 $\dmod{p_{ct}}$ &\dmod{3D Centroid} Point \dadd{described in the local frame}\\
 % $ap_{num}$ &Number of Candidate Appearances\\
 % $t_{last\_det}$ &Last Time of Candidate Detection \\
 $\dmod{num_{pt}}$ &Number of points in cluster \dmod{$\mathcal{C}_k$} \\
 $\dmod{occ_{pct}}$ &Percentage of 2D occupation on NDVI frame $\Pi$ \\
 $\bar{u}_{ndvi}$ &Average NDVI index \\
  $\dmod{conv_{area},~conv_{vol}}$ &Convex Hull Area and Volume \dadd{in 3D}\\ \bottomrule
\end{tabular}
\end{center}
\vspace{-12pt}
\end{table}
\fi

\subsection{Tree Landmarks Correspondence Matching}\label{ssec:matching}

To achieve robust landmark prediction and association while surveying, we develop a probability calculation approach during the Kalman innovation step. 
The main task is to associate newly retrieved clusters $\mathcal{C}_k$ with landmarks $\mathcal{T}_j$ stored in memory. 
We first employ the control-input matrix $\mathbf{B}_j =
\begin{bmatrix}
\cos(\theta)\Delta t & 0;&
\sin(\theta)\Delta t & 0 \\
\end{bmatrix}$ in the prediction step (line 11), and then obtain the residual innovation $\mathbf{i}_j$ for each tree $\mathcal{T}_j$ in memory and normalize it to reflect the percentage change (line 12).

Then, we compute the Mahalanobis distance $d_\mathcal{M}$ to determine the divergence of the innovation (error) residual from zero. 
The exponential distribution function is used to transform distance values into a probability $^{\boldsymbol{k}}\boldsymbol{p}_j\in[0,1]$ %within the closed interval $[0,1]$ 
(line 14). 
Once the tree association probabilities for each cluster %$^{\boldsymbol{k}}\boldsymbol{p}_j$, of the cluster $\mathcal{C}_k$ and the trees $\mathcal{T}_j$, 
have been computed, we obtain the cluster with the closest match and index it as $max_j$. 
To increase tree association robustness, we also compute the entropy $\mathcal{H}$ across the normalized probabilities of all landmark pairs (line 16). 
This metric can quantify the level of uncertainty in landmark-to-cluster associations and help decide if a new landmark instance that matches the characteristics of a specific $\mathcal{C}_k$ should be created and added in memory. 
When the cluster complies with both probability and entropy criteria, it is considered a match with $\mathcal{T}_{max_{j}}$ (lines 17-18).
The cut-off values for these two criteria may need to be adjusted as the application domain varies; in this work, we determined them empirically via preliminary testing (i.e. $th_p:=0.5, th_\mathcal{H}:=0.8$). 
Given the computed pairs $(\mathcal{T}_j,C_k)$, we discard any matching duplicates (i.e. any potential instances of equal probability) %with respect to their probability $^{\boldsymbol{k}}\boldsymbol{p}_j$ 
and proceed with updating characteristics of interest (herein, tree width and height; see below), state estimates, and covariance matrices (lines 21-23 in Algorithm~\ref{alg:treedetectionalgorithm}). 

Since each cluster is expressed in the LiDAR frame $\mathcal{P}$, we use the inverse of the transformation matrix $_{\mathcal{P}}^{\mathcal{G}}T$ to transform each $\mathcal{T}_j$ to the global Cartesian frame created by the UTM datum. Thus, we find the closest $\mathcal{GT}_h$ candidate by computing the $\ell^2$-norm in $x-y$ plane (line 25), and as ($\mathcal{T}_j$, $\mathcal{GT}_h$) pairs are obtained the corresponding global georeferenced trees are updated (line 26). 
To handle the (likely) event of operating in GNSS-denied areas~\cite{rovira2015gnssagriculturerobots, pini2020gnssrtktest}, our developed system preserves and updates the local $\mathcal{T}_j$ map during navigation, associating it with $\mathcal{GT}_h$ during periods with minimal GNSS interference.

\vspace{-1pt}
\subsection{Tree Characteristics Estimation: Height and Width}\label{ssec:geom_traits}

\begin{figure}[!t]
\vspace{5pt}
     \centering
     \begin{subfigure}{.2\textwidth}
         \centering
  \includegraphics[width=4.9cm]{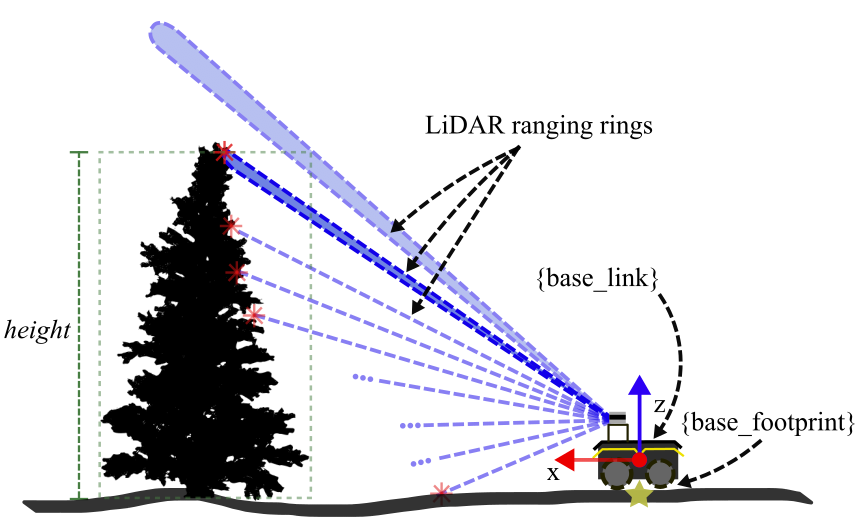}
  \vspace{-4pt}
         \caption{}
         \label{fig:jackal_height_estimation}
     \end{subfigure}\hfill
     \begin{subfigure}{.24\textwidth}
         \centering
  \includegraphics[width=4cm]{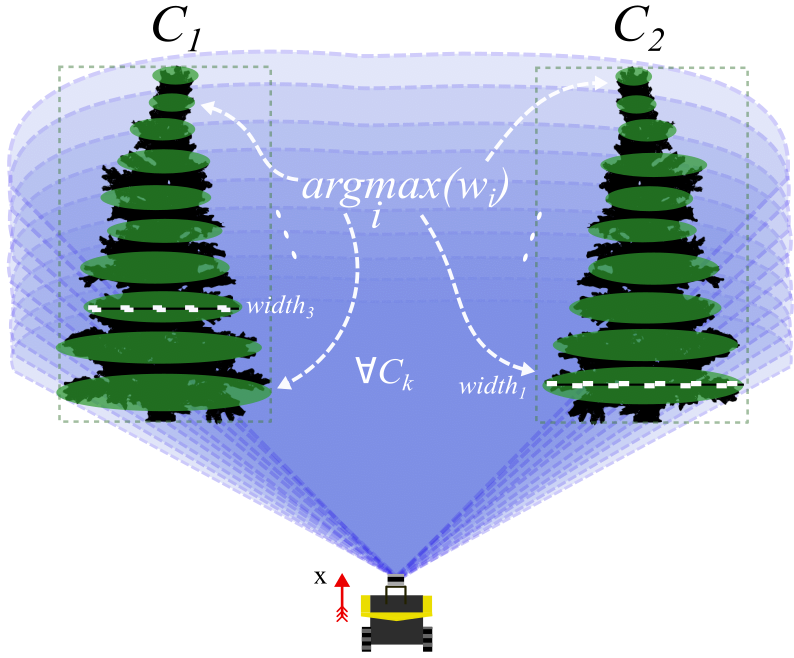}
  \vspace{-4pt}
         \caption{}
         \label{fig:jackal_width_estimation}
     \end{subfigure}
     \vspace{-4pt}
    \caption{(a) In height estimation, the robot senses a change on top $ring\_id$ detection of the cluster $\mathcal{C}_k$ and thus registers that height value with respect to $\{base\_footprint\}$. (b) In width estimation, the robot computes a tree's width value for each of the 10 $z$-slices of the clusters and stores the maximum one.}\label{fig:jackal_tree_parameter_estimation}
    \vspace{-12pt}
\end{figure}

As the robot moves in the field, the LiDAR sensor does not always fully cover the tree trunk and crown. As visualized in Figure~\ref{fig:jackal_height_estimation}, some LiDAR rays may cast on the tree's crown and others may cast away from the tree. However, as the used LiDAR has a fixed vertical angular resolution of $2^\circ$, the sensor may raycast on the tree crown's topmost part as the robot moves and hence provide information about the tree's height. Thus, to estimate the height of the tree, we keep each tree candidate's $\mathcal{T}_j$ top point's $ring\_id$ detection with the prior detections and capture the tree height on the change of the top-ranging ring. The estimated height is refined each time we detect a tree candidate; the maximum value is kept.

For width estimation, we divide each cluster in a constant number of increasing z-level point cloud batches (in our experiments we use 10) in space $\mathcal{P}$, and we seek the farthest 3D point pairs. 
Given these pairs in different z-levels, we compute the $\ell^2$-norm and we find the maximum value, which we pinpoint as the tree's diameter (i.e. width). As the robot navigates, the width of a tree candidate is recomputed for each new point cloud $\mathcal{P}$ received, and we keep the maximum value as the width of the tree candidate. Figure~\ref{fig:jackal_width_estimation} illustrates a sketch of the width estimation procedure.

\section{Experimental Testing and Validation}\label{seq:experiments}

We evaluated our proposed algorithm extensively in both realistic simulated and real-world field survey experiments. In the simulation study we performed a survey on a $2\times3$ citrus tree grid setup and assessed our algorithm's performance based on 1) using different tree placement densities, and 2) following different paths in the field. For the real-world field experiments, we deployed our robot in an orange tree grove and evaluated our algorithm's real-time performance and accuracy by following different survey paths in the orchard. 
To further enhance our evaluation results, we tested our approach on a larger citrus grove from the CitrusFarm~\cite{teng2023multimodal} open dataset. 
Ground truth information was obtained directly from the 3D tree models in the simulation cases, and from a prior synthetic map created by the authors in the real-world cases.
Comparisons against the most closely-relevant work of Jelavic $et~al.$\cite{Jelavic2021RoboticPH}  (Robotic Precision Harvesting [RPH]) are also conducted.

\begin{table*}
\vspace{6pt}
  \scriptsize
  \centering  \caption{\label{tab:simulationresults}Field Experiment Results}
  \vspace{-6pt}
\begin{tabular}{p{0.1cm}|c|cccc|cccc|p{0.2cm}|c|cc|cc}
                         \multicolumn{10}{c}{Simulated Fields}  &  \multicolumn{6}{c}{Real Fields} \\          \toprule 
\multirow{3}{*}{\hspace{-0.22cm}\shortstack{Path\\Type}} & \multirow{3}{*}{\shortstack{Tree\\Row}} & \multirow{3}{*}{\hspace{-0.1cm}\shortstack{GT\\Width\\(m)}} & \multirow{3}{*}{$\mu\pm\sigma$ (m)} & \multirow{3}{*}{\shortstack{\dmodblack{MAPE}\\Ours\\(\%)}} & \multirow{3}{*}{\shortstack{\dmodblack{MAPE}\\RPH\\(\%)}} & \multirow{3}{*}{\hspace{-0.1cm}\shortstack{GT\\Height\\(m)}} & \multirow{3}{*}{$\mu\pm\sigma$ (m)} & \multirow{3}{*}{\shortstack{\dmodblack{MAPE}\\Ours\\(\%)}} & \multirow{3}{*}{\shortstack{\dmodblack{MAPE}\\RPH\\(\%)}} & \multirow{3}{*}{\hspace{-0.13cm}\shortstack{Path\\Type}}     & \multirow{3}{*}{\shortstack{Tree\\Row}} & \multirow{3}{*}{\shortstack{\dmodblack{MAPE}\\Ours\\(\%)}} & \multirow{3}{*}{\shortstack{\dmodblack{MAPE}\\RPH\\(\%)}} & \multirow{3}{*}{\shortstack{\dmodblack{MAPE}\\Ours\\(\%)}} & \multirow{3}{*}{\shortstack{\dmodblack{MAPE}\\RPH\\(\%)}} \\
                                 &                           &                     &                       &                      &                           &                                               &                       &                                                 &                          &                                &                           &                           &                          &                           &                        \\
                                 &                           &                     &                       &                      &                           &                                               &                       &                                                 &                          &                                &                           &                           &                          &                           &                          \\   \toprule 
\multirow{3}{*}{\rotatebox[origin=c]{90}{Straight}}  & $front$                     & 2.71                & 2.65                  $\pm$ 0.08                 & \dmodblack{\textbf{0.23}}                   & 2.55                  & 2.62                & 2.58                  $\pm$ 0.09                 & 8.17                   & \dmodblack{\textbf{6.50}}                  & \multirow{3}{*}{\rotatebox[origin=c]{90}{Straight}}      & $front$                   & \dmodblack{\textbf{5.71}}                    & 10.19                 & \dmodblack{\textbf{4.69}}                   & 12.24                 \\
                           & $middle$                    & 2.67                & 2.50                  $\pm$ 0.31                 & \dmodblack{\textbf{1.93}}                   & 2.78                  & 2.74                & 2.44                  $\pm$ 0.16                 & \dmodblack{\textbf{4.00}}                   & 10.27                 &                                & $middle$                    & \dmodblack{\textbf{0.24}}                    & 6.99                  & \dmodblack{\textbf{9.74}}                   & 11.46                 \\
                           & $back$                      & 2.71                & 2.43                  $\pm$ 0.31                 & \dmodblack{\textbf{0.17}}                   & 4.07                  & \dmodblack{2.68}               & 2.26                 $\pm$ 0.29                 & \dmodblack{\textbf{2.39}}                   & 8.74                  &                                & $back$                    & \dmodblack{\textbf{11.94}}                   & 13.43                 & 15.20                  & \dmodblack{\textbf{7.35}}                  \\ \midrule
\multirow{3}{*}{\rotatebox[origin=c]{90}{$S$-type}}    & $front$                     & 2.71                & 2.98                  $\pm$ 0.36                 & \dmodblack{\textbf{0.29}}                   & 1.05                  & 2.62                & 3.55                  $\pm$ 1.48                 & \dmodblack{\textbf{3.03}}                   & 6.83                  & \multirow{3}{*}{\rotatebox[origin=c]{90}{$S$-type}}        & $front$                    & \dmodblack{\textbf{2.58}}                    & 9.78                  & \dmodblack{\textbf{10.82}}                  & 12.86                 \\
                           & $middle$                    & 2.67                & 2.59                  $\pm$ 0.12                 & 12.14                  & \dmodblack{\textbf{10.49}}                 & 2.74                & 2.95                  $\pm$ 1.07                 & 14.71                  & \dmodblack{\textbf{11.25}}                 &                                & $middle$                     & \dmodblack{\textbf{2.60}}                    & 5.89                  & \dmodblack{\textbf{6.34}}                   & 14.39                 \\ 
                           & $back$                      & 2.71                & 2.70                  $\pm$ 0.39                 & \dmodblack{\textbf{0.37}}                   & 2.19                  & \dmodblack{2.68}               & 2.65                  $\pm$ 0.47                 & \dmodblack{\textbf{7.45}}                   & 10.32                 &                                & $back$                   & \dmodblack{\textbf{11.79}}                   & 13.43                 & 12.25                  & \dmodblack{\textbf{6.62}}                  \\ \midrule
\multirow{3}{*}{\rotatebox[origin=c]{90}{$N$-type}}   & $front$                     & 2.71                & 2.99                  $\pm$ 0.56                 & \dmodblack{\textbf{5.75}}                   & 12.04                 & 2.62                & 3.38                  $\pm$ 1.16                 & \dmodblack{\textbf{6.50}}                   & 7.30                  & \multirow{3}{*}{\rotatebox[origin=c]{90}{$N$-type}}        & $front$                    & \dmodblack{\textbf{5.30}}                    & 15.63                 & \dmodblack{\textbf{8.57}}                   & 10.00                 \\
                           & $middle$                    & 2.67                & 2.91                  $\pm$ 0.34                 & \dmodblack{\textbf{7.57}}                   & 23.45                 & 2.74                & 3.01                  $\pm$ 1.01                 & \dmodblack{\textbf{1.12}}                   & 15.33                 &                                & $middle$                     & \dmodblack{\textbf{3.56}}                    & 31.37                 & \dmodblack{\textbf{9.02}}                   & 11.95                 \\
                           & $back$                      & 2.71                & 2.96                  $\pm$ 0.13                 & \dmodblack{\textbf{11.46}}                  & 13.35                 & \dmodblack{2.68}               & 2.90                 $\pm$ 1.00                 & \dmodblack{\textbf{1.76}}                   & 4.09                  &                                & $back$                    & \dmodblack{\textbf{16.57}}                  & 35.67                 & \dmodblack{\textbf{9.07}}                   & 19.12                 \\   
  \midrule
\multirow{3}{*}{\rotatebox[origin=c]{90}{$4\times 4$}}       & $front$                     & 2.71                & 2.63                  $\pm$ 0.09                 & \dmodblack{\textbf{0.30}}                   & 4.51                  & 2.58                & 2.58                  $\pm$ 0.08                 & \dmodblack{\textbf{10.49}}                  & 10.97                 &                                &                           &                           &                          &                           &                          \\
                           & $middle$                    & 2.67                & 2.51                  $\pm$ 0.43                 & \dmodblack{\textbf{10.02}}                  & 10.59                 & 2.62                & 2.31                  $\pm$ 0.30                 & 11.85                  & \dmodblack{\textbf{8.80}}                  & \multirow{5}{*}{\rotatebox[origin=c]{90}{CertFarm}} & $row\_1$                      & 4.77                   & \dmodblack{\textbf{3.71}}                  & \dmodblack{\textbf{9.53}}                   & 13.28                 \\
                           & $back$                      & 2.77                & 2.82                  $\pm$ 0.27                 & \dmodblack{\textbf{5.21}}                   & 6.99                  & \dmodblack{2.67}               & 2.42                  $\pm$ 0.36                 & \dmodblack{\textbf{5.16}}                   & 9.50                  &                                & $row\_2$                      & \dmodblack{\textbf{2.51}}                   & 4.93                  & \dmodblack{\textbf{4.26}}                   & 11.91                 \\ \cmidrule[0.05pt](lr){1-10} 
\multirow{3}{*}{\rotatebox[origin=c]{90}{$6\times6$}}       & $front$                     & 2.71                & 2.67                  $\pm$ 0.06                 & \dmodblack{\textbf{0.22}}                   & 2.64                  & 2.58                & 2.58                  $\pm$ 0.08                 & \dmodblack{\textbf{8.72}}                   & 11.87                 &                                & $row\_3$                      & \dmodblack{\textbf{5.26}}                   & 7.34                  & 9.95                   & \dmodblack{\textbf{6.95}}                  \\
                           & $middle$                    & 2.67                & 2.52                  $\pm$ 0.27                 & \dmodblack{\textbf{1.36}}                   & 2.25                  & 2.62                & 2.45                  $\pm$ 0.19                 & 8.59                   & \dmodblack{\textbf{6.36}}                  &                                & $row\_4$                      & \dmodblack{\textbf{10.97}}                  & 23.88                 & \dmodblack{\textbf{7.08}}                   & 12.15                 \\
                           & $back$                      & 2.77                & 2.44                  $\pm$ 0.28                 & \dmodblack{\textbf{3.64}}                   & 4.50                  & \dmodblack{2.67}               & 2.19                   $\pm$ 0.48                 & \dmodblack{\textbf{4.30}}                   & 12.26                 &                                & $row\_5$                      & \dmodblack{\textbf{12.64}}                  & 16.32                 & \dmodblack{\textbf{3.02}}                   & 3.93         \\ \bottomrule       
\end{tabular}
\vspace{-15pt}
\end{table*}

\vspace{-3pt}

\subsection{Evaluation in Simulated Tests}

Simulated tests were conducted within the Gazebo simulator. %~\cite{koenig2004design}. 
The mobile robot model was modified to integrate the actual sensors' configuration and noise models. To increase the fidelity of the robot's traversability with respect to the real world, we created a custom 3D world, generated by aerial imagery data of a citrus field at UCR's Agricultural Experimental Station (AES; $33^\circ{}~58'~3.2592''~N,~
117^\circ{} 20'~7.0296'' W$) with the Agisoft Metashape software. Various realistic citrus (lemon and orange) tree instances were created by the Helios library~\cite{helios} and integrated into the simulated world. 
Figure~\ref{fig:gazebo_simulated_world} depicts the robot traversing the simulated field. 
In all simulated tests, the robot had constant linear ($v_x=0.5~m/sec$) and angular ($\omega_{\theta}=1~rad/sec$) velocities.

\paragraph*{\bf Using different tree placement densities} %\label{seq:simulated_experiment_grid} 
We generated three different world models by having the citrus trees $4$, $5$, and $6$ $m$ away from their neighboring trees. These grids are denoted $\{3,4,5\}~m$ width $\times$ $\{3,4,5\}~m$ length, respectively. The instance of $5~m$ width $\times$ $5~m$ length is depicted in Figure~\ref{fig:gazebo_simulated_world}. These selections were made so as to be consistent with tree placement in commercial citrus orchards. 
From an algorithmic standpoint, they enable us to examine our method's performance in both sparsely- and densely-planted orchards. Table~\ref{tab:simulationresults} presents the evaluation results of \dch{five} independent trials in each of the three tree-grid cases.

\begin{figure}[!t]
    \vspace{6pt}
     \centering
     \hspace{-12pt}
     \begin{subfigure}{.15\textwidth}
         \centering
  \includegraphics[trim={0cm 0cm 0.90cm 1.5cm},clip,width=2.9cm]{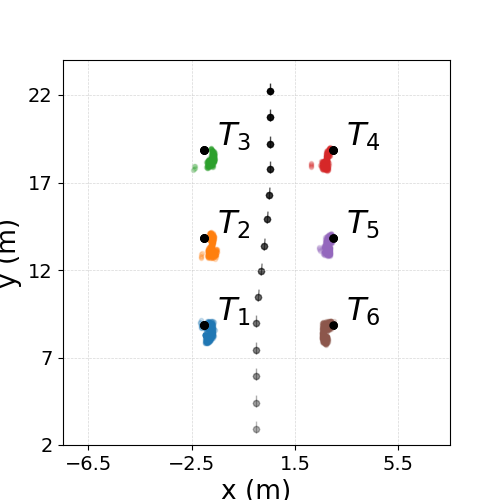}
  \vspace{-15pt}
         \caption{}
         \label{fig:panoramic_simulation_straight}
     \end{subfigure}\hspace{2pt}
     \begin{subfigure}{.15\textwidth}
         \centering
  \includegraphics[trim={0cm 0cm 0.90cm 1.5cm},clip,width=2.9cm]{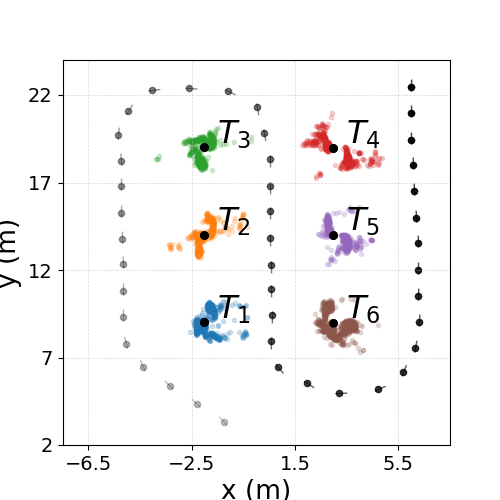}
   \vspace{-15pt}
         \caption{}
         \label{fig:panoramic_simulation_n}
     \end{subfigure}\hspace{2pt}
     \begin{subfigure}{.15\textwidth}
         \centering
  \includegraphics[trim={0cm 0cm 0.90cm 1.5cm},clip,width=2.9cm]{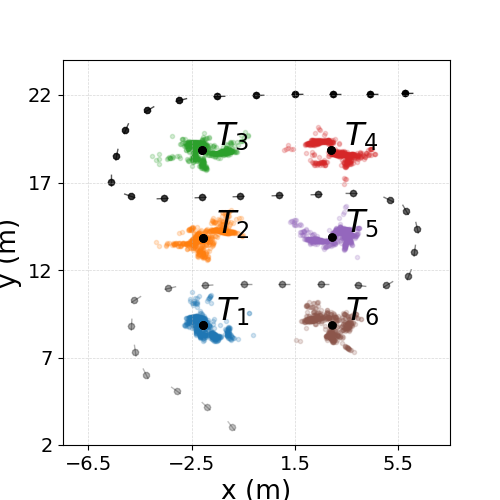}
   \vspace{-15pt}
         \caption{}
         \label{fig:panoramic_simulation_s}
     \end{subfigure}
     \vspace{-3pt}
    \caption{Panoramic views of the simulated tests for (a) straight-line, (b) $N$-type, and (c) $S$-type robot trajectories on the $5~m$ width $\times$ $5~m$ length tree grid setup. \dch{Along with the sampled and decreasingly transparent (with respect to time)} robot poses, there is a colorized illustration of the detected tree candidates' $p_{centroid}$ points, which are projected on the UTM $x-y$ plane to demonstrate their distribution around the groundtruth $\mathcal{GT}$ tree positions.}\label{fig:simulated_panoramic_plots}
    \vspace{-18pt}
\end{figure}

To perform the simulated surveys, we provided a goal waypoint in all grid cases for the robot to follow. The goal was set $23~m$ ahead from the origin along the longer side of the field (see Figure~\ref{fig:panoramic_simulation_straight}). The robot does not stop at any point unless it has reached the goal.

Runtime comparisons between our method and RPH indicated a twofold time increase in tree detection in the latter case. RPH requires on average $0.53~sec$ for detecting the clusters in point clouds of $27$-$30K$ points, whereas our method has $0.22~sec$ on average detection runtime in the same point clouds and can provide up to 2x more tree detections in the same time frame. 
One of the main reasons causing delays in RPH seems to be the PCA step for locating clusters with z-axis divergence. 
Tree detection accuracy was similar to our $S$-case as both methods use Euclidean clustering, as a basis, to form the clusters $\mathcal{C}_k$.

\begin{figure*}[!t]
\vspace{6pt}
     \centering
     \begin{subfigure}{.33\textwidth}
         \centering
  \includegraphics[height=3.05cm]{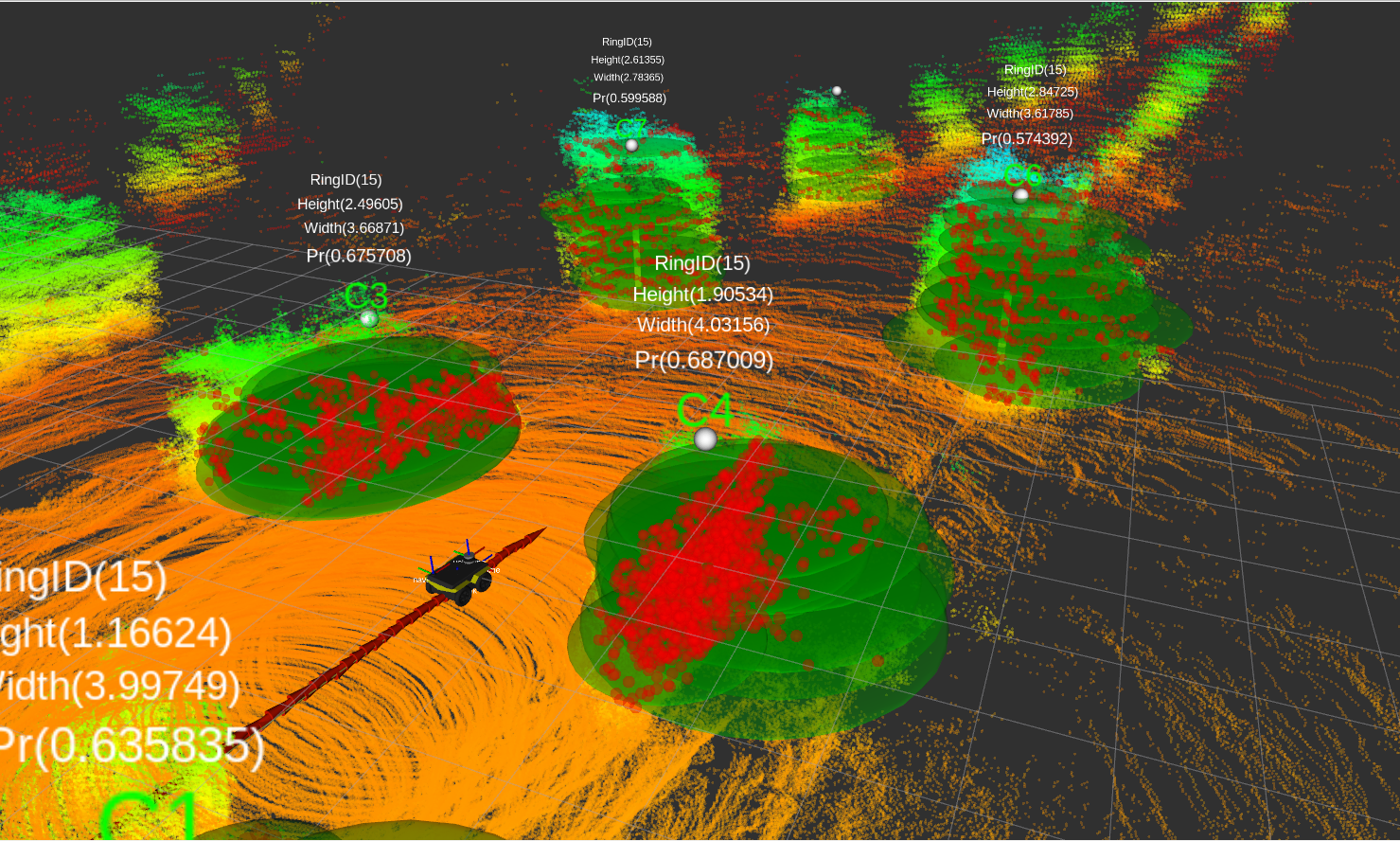}  
         \caption{}
         \label{fig:n_type_real_panoramic}
     \end{subfigure}\hfill
     \begin{subfigure}{.33\textwidth}
         \centering
  \includegraphics[height=3.05cm, trim={0 0 1cm 0},clip]{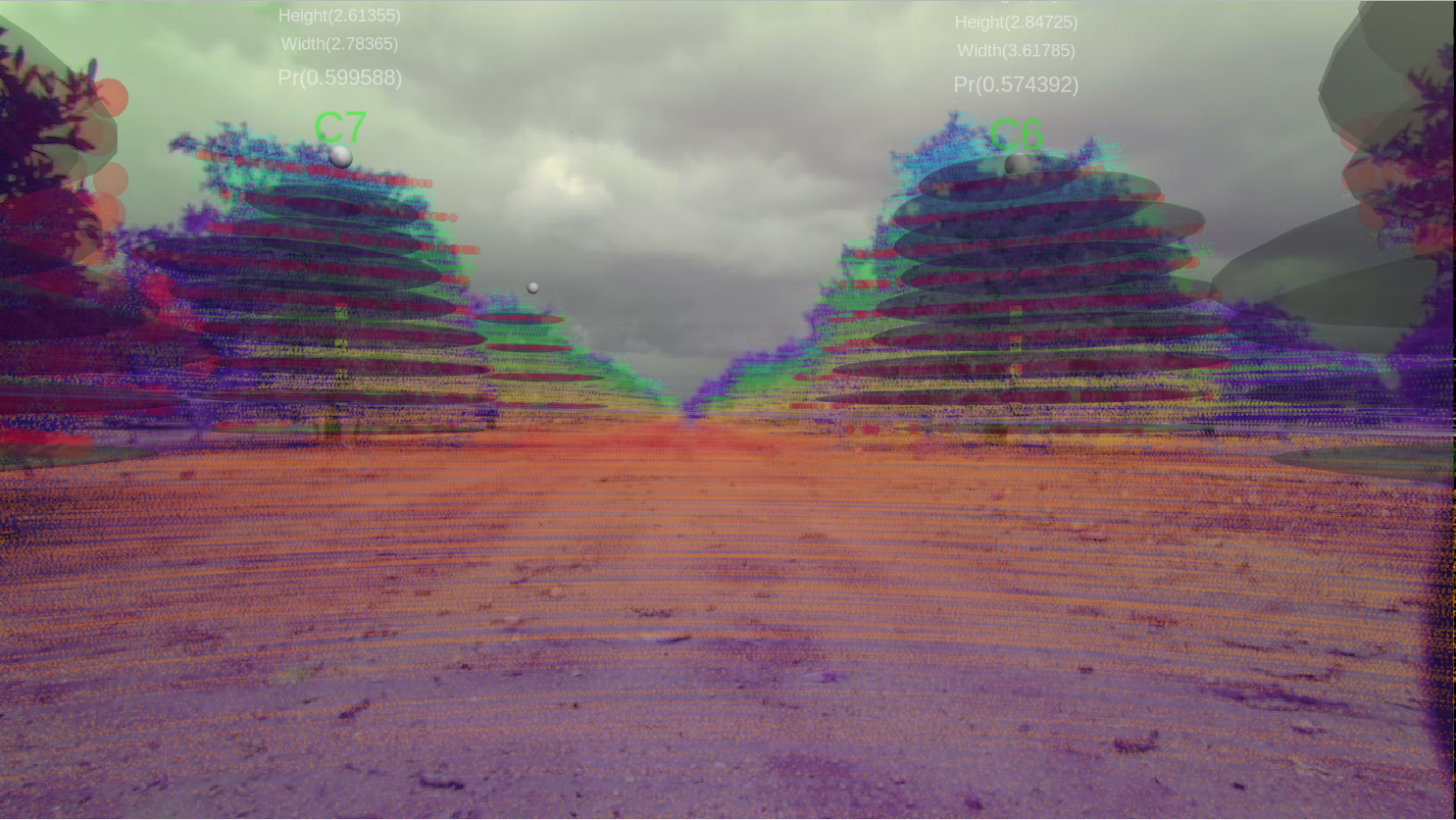}
         \caption{}
         \label{fig:n_type_real_rgn}
     \end{subfigure}\hfill
     \begin{subfigure}{.33\textwidth}
         \centering
\includegraphics[height=3.05cm]{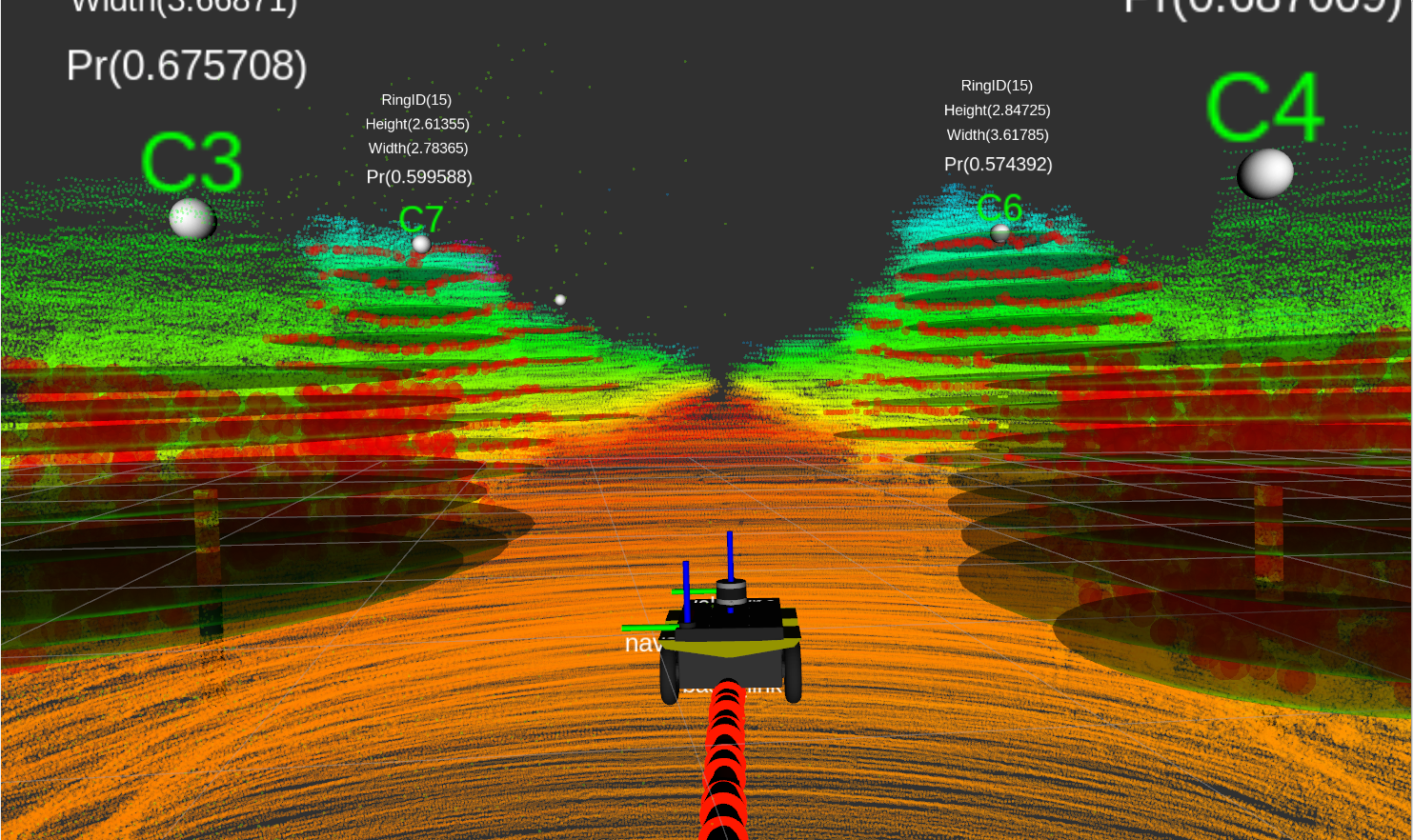}
         \caption{}
         \label{fig:n_type_real_pov}
     \end{subfigure}
     \vspace{-6pt}
    \caption{Snapshot from the $N$-type field experiment in the orchard. (a) Tree detection status along with detected \dmod{$\mathcal{C}_k$} tree clusters. The image depicts information stats of each detected and matched cluster during that time (green rings  are used for trees' width estimation, and red points are part of each cluster). (b) The corresponding RGN frame along with the projected LiDAR information and depicted tree information. (c) A closer third-person view of the Jackal in RViz closely matching the RGN frame. (Figure best viewed in color.)}
    \label{fig:n_type_real}
    \vspace{-20pt}
\end{figure*}

Results from tree characteristics extraction (rows of Path Type: ``Straight" in Table~\ref{tab:simulationresults}) show that our algorithm enables the robot to perform both tree width and height estimation very accurately in all cases. The maximum Mean Absolute Percentage (MAP) errors observed were $12.1\%$ and $14.7\%$ for width and height estimations, respectively, over all tree configurations. 
Since the robot starts from point $(0,0)$, it captures more observations (greater view) of the $front$ row, and thus achieved $2\%\mbox{-}5\%$ of absolute percentage error less compared to those for the $middle$ and $back$ row. 
On the other hand, height estimation error decreased by $3\%$ in more sparse fields, namely cases of $5\times 5$ and $6\times 6$ because of the broader field of view of the robot. The RPH method had up to $3\%$ higher average MAP error in both width and height estimation, which can be linked to the slower runtime and thus a smaller number of tree observations. 
In our approach, comparably higher errors (on average) can be linked to the fact that the algorithm is using ranging information from the topmost part of a tree to estimate its height, but the former may be assessed multiple times during robot traversal thus including measurement variability.

\paragraph*{\bf Following different paths in the field} The second set of simulated tests examined the behavior of the algorithm as the followed path may vary. We used the $5\times 5$ tree placement grid, and tested our algorithm's performance in more complex path types, namely $N$-type and $S$-type paths, in comparison to the aforementioned straight-line findings (first three rows of Table~\ref{tab:simulationresults}). Sample $N$-type and $S$-type paths are shown in Figure~\ref{fig:simulated_panoramic_plots}.

From the reported results, we observe that our algorithm scores $4.3\%$ and $8.3\%$ of MAP error in width estimation and $8.4\%$ and $3.1\%$ of MAP error in height estimation for $S$-type and $N$-type trajectories, respectively. \dch{One of the main factors contributing to the difference in accuracy compared with the straight case, is that the robot performs closer maneuvers inside the field}. \dch{Similar performance is also notable in the RPH case, with an increase up to $8\%$ and $5\%$ of average MAP error in width and height estimation, respectively. In both systems, $N$-type surveys led to better estimations of the height level of the trees due to the longer straight lines within the field, while $S$-type surveys helped with a better estimation of the tree width due to the multiple traverses closer to the tree rows.} In all, despite the different maneuvers in the field, our algorithm had an average of $5.3\%$ in MAP error in both of these cases, demonstrating its robustness in width and height estimation.

\subsection{Evaluation and Validation in Field Experiments}
We evaluated our algorithm to validate its efficacy in field experiments. We performed a real survey in a citrus tree grove at UCR AES. We focused on a $2\times3$ tree grid part of the selected field (to be consistent with the simulation test setup). This grid featured $6.5~m$ width $\times$ $7~m$ length tree placement, which is a setup not included in simulated tests. On the experiment day, the reported wind speed was at $16~km/h$, with gusts up to $40~km/h$.

Quantitative results for all cases are provided in Table~\ref{tab:simulationresults}, whereas Figure~\ref{fig:n_type_real} \dch{visualizes} instances of the algorithm's behavior. 
Despite the windy conditions in the field, the robot scored a maximum of $16.6\%$ and $15.2\%$ MAP error in tree width and height estimation, respectively, over all trajectory types. 
Consistent with simulation findings, longer in length and more straight paths led to better results as observed by the straight case scoring on average $3\%$ and $4\%$ less MAP errors, compared with $S$ and $N$ cases, respectively. \dch{Similarly, on the RPH case, the height estimation results appear to be similar in all cases, averaging at $2.5\%$ MAP more than ours. However, in width estimation on RPH case, the $N$ case was the most noisy, as the windy conditions along with the restricted runtime of the algorithm, are not ideal for width estimation (as seen also in the simulated experiments), by scoring $10\%$ of additional average MAP error.}
Overall, our algorithm gave less than $17\%$ MAP error of geometric trait estimation, over all tested cases, even at challenging field conditions (high wind and diverse ambient light variations).  

Additionally, in these cases, the robot collected the average NDVI measurements from each tree. This information is integral into our landmark association system but at the same time it can be ``overloaded" to help monitor the identified trees' health (which is currently the use of NDVI field data). For the cases of the $N$ and $S$-typed trajectories (since they cover all the sides of the inspected trees), we observed resulting scores of $80\%$ of an average relative distance, by noting $+0.55$, $+0.61$, and $+0.62$ of an average NDVI index, for the $front$, $middle$, and $back$ row of the inspected trees.

To further support our evaluation step, we tested on a multi-modal dataset from CitrusFarm, that includes RGN and 3D LiDAR modalities collected by a mobile robot. We selected a survey path with $N$-type trajectory, in a $5\times3$ citrus tree grove of $29\times 21~m^2$ area size. With reference to Table~\ref{tab:simulationresults}, our system performs well by having maximum MAP error at $12.7\%$ at width estimation and $9.9\%$ at height estimation, respectively, showcasing its scalability into larger fields. It is worth noting that, there was no reported wind in the used dataset of CitrusFarm as it was captured on a sunny day, which contributes to the observed improvements on average MAP errors. 
The RPH approach showcased an additional $3.4\%$ of average MAP error in both estimation cases.

\vspace{-1pt}
\section{Conclusions}\label{seq:conclusions}
\vspace{-1pt}

In this work, we introduced an algorithm for real-time and on-the-go tree detection and geometric traits estimation with wheeled mobile robots using onboard sensors.
At its core, the algorithm uses diverse multi-modal sensory data (domain-specific RGN images and 3D point clouds) for tree detection, without relying on any specific tree parts.  
Key to this process is our proposed multivariate and entropy-based 
approach, based on an underlying Kalman Filter, which helps obtain local tree associations, calculate their physical and vegetation-based characteristics, and assign/correlate them with global georeferenced trees for inspection updates. 
Extensive evaluation in both %realistic 
simulation tests and physical field experiments in an orchard demonstrated the efficacy of our method
and real-time robustness. 
Future research directions enabled by this work include extending to multi-robot systems for larger field coverage and employing multi-modal sensing for in-field and real-time analysis of tree geometry, health, and crop yield indices.


%%%%%%%%%%%%%%%%%%%%%%%%%%%%%%%%%%%%%%%%%%%%%%%%%%%%%%%%%%%%%%%%%%%%%%%%%%%%%%%%

%%%%%%%%%%%%%%%%%%%%%%%%%%%%%%%%%%%%%%%%%%%%%%%%%%%%%%%%%%%%%%%%%%%%%%%%%%%%%%%%

%%%%%%%%%%%%%%%%%%%%%%%%%%%%%%%%%%%%%%%%%%%%%%%%%%%%%%%%%%%%%%%%%%%%%%%%%%%%%%%%
%\section*{APPENDIX}


%\section*{ACKNOWLEDGMENT}

%%%%%%%%%%%%%%%%%%%%%%%%%%%%%%%%%%%%%%%%%%%%%%%%%%%%%%%%%%%%%%%%%%%%%%%%%%%%%%%%

\bibliographystyle{ieeetr} 
\bibliography{ref}

\end{document}